\documentclass[journal]{IEEEtran}
\IEEEoverridecommandlockouts
\usepackage{cite}
\usepackage{amsmath,amssymb,amsfonts}
\usepackage{algorithmic}
\usepackage{graphicx}
\usepackage{textcomp}
\usepackage{booktabs}
\usepackage{mathtools}
\usepackage{bm}

\begin{document}

\title{Surf-Style: Surface-Based Scene Stylization}

\author{Wanrong Hong, Ancheng Lin, Mukesh Prasad, and Jun Li%
\thanks{This work has been submitted to the IEEE for possible publication. Copyright may be transferred without notice, after which this version may no longer be accessible.}%
\thanks{The authors are with the School of Computer Science, Australian Artificial Intelligence Institute (AAII), University of Technology Sydney, Sydney 2007, Australia.}%
\thanks{Corresponding author: Jun Li (e-mail: jun.li@uts.edu.au).}}

\markboth{}{Hong \MakeLowercase{\textit{et al.}}: Surf-Style: Surface-Based Scene Stylization}

\maketitle

\begin{abstract}
Consistent and controllable 3D stylization is critical for visual computing applications such as immersive visualization, augmented and virtual reality content creation, and artistic scene rendering.
While 2D image stylization has advanced significantly, extending these methods to 3D remains challenging due to the need for cross-view consistency, where stylized renderings must align across all viewpoints. 
Applying 2D stylization independently to each rendered view often results in inconsistencies across perspectives.
Existing methods attempt to improve consistency by leveraging implicit radiance field representations, which model scene appearance as a function of spatial locations. 
However, these methods struggle with ambiguity in updating colors, as the indirect dependence between pixel appearance and learnable scene parameters makes it difficult to enforce consistent stylization during optimization.
This work presents Surface-based Scene Stylization (Surf-Style), which reconstructs object surfaces in a 3D scene and uses these surfaces to achieve geometric consistency and localized stylization.
Surf-Style defines a learnable appearance model over 3D space, which is modified to replicate the visual style of a reference image.
Compared to state-of-the-art methods, our method achieves superior consistency and offers greater flexibility in controllable stylization, including the ability to target specific surface regions and combine different styles.
\end{abstract}


\section{Introduction}
\label{sec:introduction}
\IEEEPARstart{C}{hanging} the style of visual signals is widely used in digital art, advertising, and game graphics.
Early methods \cite{gatys2016image, isola2017image} apply style transfer to 2D images. These models learn from example images of a given style and generate new images by minimizing perceptual metrics between the output and the reference style.
However, they do not account for the 3D geometry of the scene.
As a result, when applied to practical applications such as streaming stylized scenes or rendering videos, these methods struggle to maintain consistency across different viewpoints.

\begin{figure}[t]
    \centering
    \includegraphics[width=0.9\columnwidth]{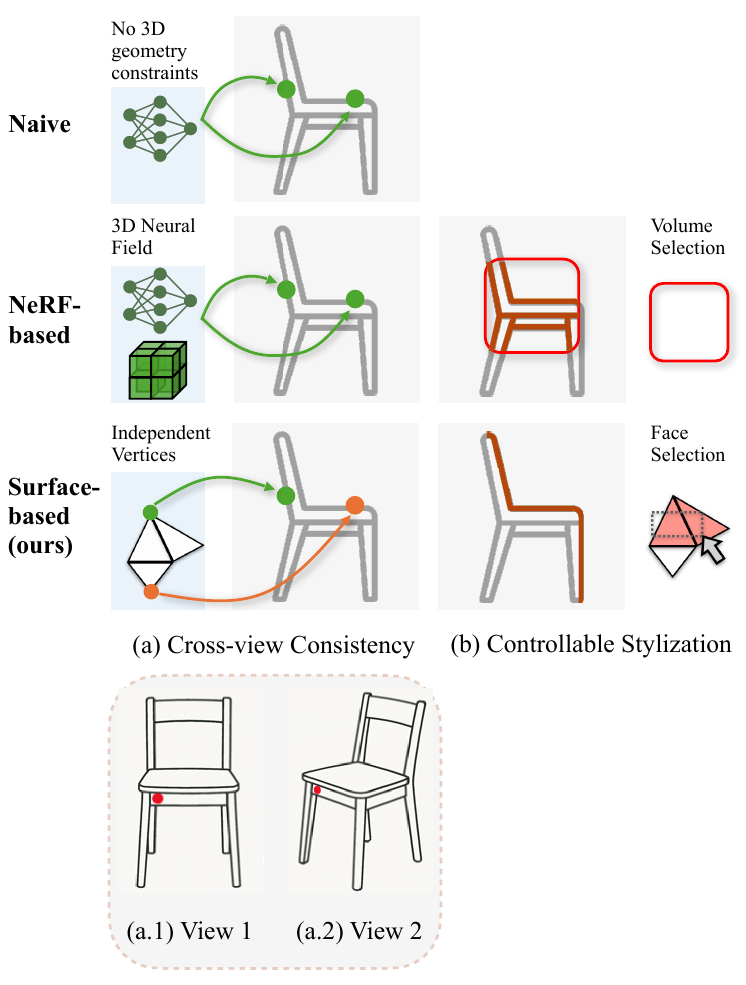} \\
    \caption{Difference between naive image-based stylization, NeRF-based
    methods, and our surface-based approach in 3D scene stylization. \textbf{(a)
    Cross-view Consistency}: Naively stylizing rendered images without enforcing
    3D geometry constraints lead to cross-view inconsistencies. NeRF-based
    methods \cite{chen2024upst, huang2022stylizednerf, zhang2022arf,
    zhang2023refnpr} offer limited consistency, as they use shared parameters
    (e.g., neural networks, voxel grids) to model appearance within a volume,
    causing appearance changes to spill over into unrelated regions. In
    contrast, our surface-based method recovers a surface model from the trained
    NeRF and stores appearance information independently at face vertices,
    reducing conflicts and inconsistencies. \textbf{(a.1)} and \textbf{(a.2)} 
    Cross-view consistency illustration. \textbf{(b) Controllable stylization}: 
    The naive method does not support geometry-aware selection.
    NeRF-based approaches struggle to target specific regions, as they do not
    explicitly represent object structures. Our surface-based method enables
    selection over individual faces through explicit geometric modeling,
    allowing for localized stylization.
    }
    \label{fig:volume_surface_compare}
\end{figure}

As illustrated in Fig.~\ref{fig:volume_surface_compare} (a.1) and (a.2), a red point on the surface appears at different locations in the image depending on the viewpoint.
Since both observations correspond to the same point on the same 3D surface, their stylized appearance should remain consistent across all views.
To achieve such invariance, a stylization model is required to understand the 3D geometry of the scene.

Recent novel-view stylization methods \cite{chen2024upst, huang2022stylizednerf, zhang2022arf, zhang2023refnpr, radl2023laenerf, cheng2023AGAP} have adopted neural radiance fields (NeRFs) for 3D stylization.
These neural fields provide access to underlying 3D geometry, enabling models to incorporate spatial information into the stylization process.
However, NeRFs rely on volumetric rendering, where each pixel color is computed by integrating contributions from multiple sampled points along a ray.
This indirect dependence between pixel appearance and learnable scene parameters makes it difficult to enforce consistent stylization during optimization.

Additionally, existing methods lack precise control over which areas undergo style changes, meaning they cannot confine stylization to user-specified surface.
Recent works \cite{zhang2023refnpr, radl2023laenerf, cheng2023AGAP} attempt to address this issue by using grids to define specific regions for modification. 
However, these approaches rely on axis-aligned selections, leading to unintended modifications outside the target area.
\cite{StylizedGS} has explored language-guided region selection for controllable stylization in 3D Gaussian Splatting representations.
An alternative interactive approach \cite{radl2023laenerf} allows users to initiate region selection by clicking on a voxel, then expands the region iteratively in all directions based on voxel occupancy and spatial proximity. However, this volumetric expansion lacks surface awareness, making it less suitable for selecting specific surfaces.

To address these challenges, we propose a framework named the ``Surface-based Scene Stylization Model (Surf-Style),'' which reconstructs object surfaces in a 3D scene and uses these surfaces to achieve geometric consistency and localized stylization.
Surf-Style defines a learnable appearance model over 3D space, which is modified to replicate the visual style of a reference image.
During rendering, the model projects the appearance-modified surfaces onto the image plane.
Because the appearance model is defined in 3D space and remains invariant to camera pose, the resulting stylization is consistent across different viewpoints.
The appearance model includes both view-independent diffuse components and view-dependent specular components. The diffuse color is optimized to reflect the desired style, while the specular features, which are responsible for producing view-dependent effects, are kept fixed to preserve appearance consistency across views.
Specifically, we use mesh as the representation of the surface. The diffuse and specular terms are attached to mesh vertices, which enable the localized stylization by selecting the desired regions on the mesh.

Experiments show that Surf-Style achieves superior geometric consistency
compared to state-of-the-art methods on both synthetic and real-world datasets.
In addition, it enables users to apply styles to specific regions
and control the styles of different parts or objects separately.
In summary, this paper presents two key contributions:
\begin{itemize}
    \item We propose the Surf-Style framework, combining geometry extraction and
    stylization processes to achieve geometric consistency and detailed control
    in 3D stylization.
    \item Our method supports localized stylization by attaching view-independent and view-dependent appearance features to mesh vertices, allowing flexible control over the style applied to different regions of the surface.
\end{itemize}

\section{Related Work}

\subsection{Neural Radiance Fields}

Novel view synthesis is generating a new, unseen view of a 3D object or scene from one or more existing views. 
The research on Neural Radiance Fields (NeRFs) \cite{mildenhall2021nerf} leads to progress in novel view synthesis, as well as its extensions, such as scene editing \cite{yuan2022editing}, dynamic scene rendering \cite{pumarola2021dnerf}, and scene reconstruction \cite{Tang2023NeRF2Mesh, Wei2023NeuManifoldNW}. 
A major challenge in NeRFs is the time-consuming training and inference process.
DVGO \cite{Sun22} utilizes separate voxel grids for density and features, employing post-activation interpolation to hasten training.
Plenoxels \cite{Fridovich2022} caches a pre-trained NeRF representation using an octree structure, employing spherical harmonic (SH) coefficients for view-dependent color calculation.
Tensor Field Networks (TensoRF) \cite{Chen22TensoRF} represents a radiance field as a 4D tensor, using vectors and matrices to factorize grid features' appearance and geometry into low-rank tensors.
Instant-NGP \cite{muller2022instant} introduces multiresolution hash encoding to speed up the training and evaluation of neural graphics primitives without compromising quality. 
Another challenge in NeRFs is accurately capturing details at varying scales while addressing aliasing artifacts. 
Mip-NeRF \cite{Barron_mipnerf} addresses the issue of aliasing by introducing a cone-tracing mechanism that models rays as cones instead of single points. Mip-NeRF 360 \cite{Barron22_Mip360} builds on Mip-NeRF, targeting unbounded, complex scenes.
Finally, there is growing interest in dynamic NeRFs, which model scenes involving motion or deformation. Progress in this area has enabled the modeling of dynamic scenes from both synchronized multi-view videos  \cite{Liu22_DeVRF} and even single-view monocular videos \cite{pumarola2021dnerf}. 

\subsection{Surface Reconstruction from NeRFs}

Surface reconstruction in NeRFs refers to the process of extracting and representing the underlying 3D geometry of a scene. The goal is to recover accurate surfaces from the implicit volumetric representation NeRFs provide.
NeuS \cite{Wang2021} is a pioneer in this field, leveraging a hybrid representation that converts a Signed Distance Function (SDF) into a density field while utilizing NeRF's learning framework. 
VolSDF \cite{Yariv2021VolSDF} represents density using Laplace's cumulative distribution function (CDF) applied to an SDF.
UNISURF \cite{Oechsle2021UNISURF} focuses on solid object representation with 3D surfaces and view-dependent colors, establishing a unified framework for implicit occupancy and radiance fields.
These methods not only capture view-dependent appearances but also enable mesh extraction, making them well-suited for various downstream applications.
Subsequent works further refine the extracted meshes to improve geometric accuracy and texture quality \cite{Xiang2021NeuTexNT, Yang2022NeuMesh, Chen2022MobileNeRF, Wei2023NeuManifoldNW, Tang2023NeRF2Mesh}.
NeuMesh \cite{Yang2022NeuMesh} transforms MLP-based neural implicit fields into mesh-based representations by encoding the disentangled geometry and texture codes on mesh vertices.
MobileNeRF \cite{Chen2022MobileNeRF} represents NeRF using a grid mesh, enabling real-time rendering while reducing memory consumption.
NeuManifold \cite{Wei2023NeuManifoldNW} pre-trains a neural implicit field with TensoRF \cite{Chen22TensoRF} and extracts watertight manifold meshes using DMTet \cite{Shen2021DMTet} or DiffMC \cite{Wei2023NeuManifoldNW}, achieving high-quality geometry and neural textures.
NeRF2Mesh \cite{Tang2023NeRF2Mesh} introduces an iterative mesh refinement algorithm for reconstructing textured surface meshes from multi-view RGB images.
In this paper, we adopt NeRF2Mesh models for explicit mesh extraction as the geometric representation for stylization.

\subsection{3D Stylization}
Image stylization, the process of transforming an input image to accentuate a specific artistic or visual style, has been extensively explored in literature \cite{gatys2016image, isola2017image, zhu2017unpaired, Lin2021, Lee2020}. 
However, extending stylization techniques to 3D scenes presents a unique challenge: unlike the regularity of image pixels, 3D scene representations, such as meshes or point clouds, exhibit a more irregular structure, making stylization more difficult to achieve \cite{ hollein2022stylemesh, cao2020psnet, liu2018paparazzi}. 

Recently, advancements in NeRF have been applied to the field of 3D stylization. Several studies have employed hyper-networks to predict the parameters of the appearance branch within the NeRF networks \cite{chiang2022stylizing, chen2024upst}. However, hyper-networks introduce an additional layer of complexity, which can lead to increased computational and architectural demands.
ARF \cite{zhang2022arf} adopts a style loss from \cite{kolkin2022nnst}, which minimizes the distance between style features extracted from the style image and content features from the content image, aiming to achieve a close match between the stylized and reference appearance.
SNeRF \cite{nguyen2022snerf} iteratively stylizes novel views by rendering images from a pre-trained NeRF, applying a stylization module on the rendered images, and retraining the NeRF to match the stylized outputs, ultimately producing a stylized scene capable of novel view synthesis. 
Ref-NPR \cite{zhang2023refnpr} introduces a reference-based stylization framework.
Furthermore, several works have focused on accelerating the stylization process \cite{goel2022styletrf, li2024instantstyle, saroha2024gsstyle, zhang2024stylizedgs, liu2024stylegaussian} by modifying various radiance field representations \cite{Chen22TensoRF, muller2022instant, kerbl20233dgs}.
Despite significant progress in 3D stylization, challenges such as preserving geometric consistency and achieving precise control over the targeted areas for stylization remain unresolved.

\section{Method}

\begin{figure*}[t]
    \centering
    \includegraphics[width=\textwidth]{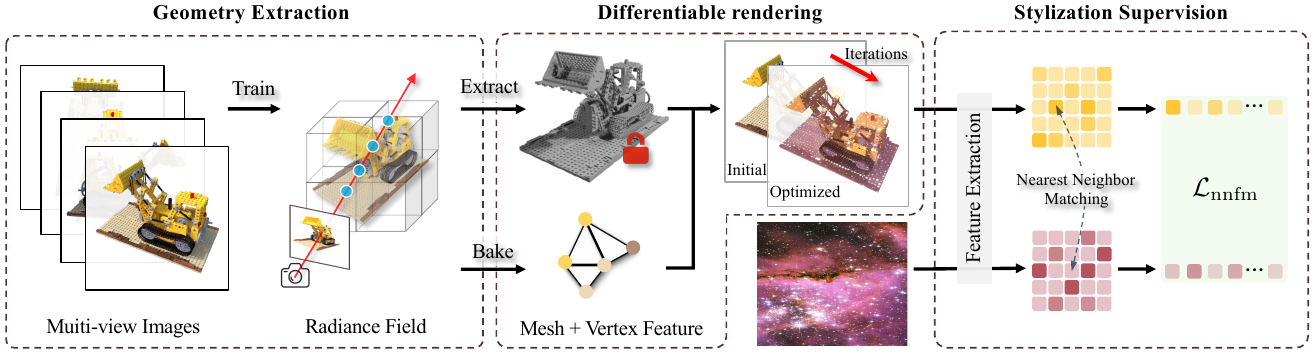}
    \caption{Workflow and main components of the proposed Surf-Style framework. The process begins with extracting explicit geometry from a radiance field trained on multi-view images, while an appearance grid captures the scene’s visual details. During stylization, the appearance information is transferred to the mesh vertices. The scene is then rendered using a custom shader within a differentiable rendering framework. A feature extraction module extracts features from both the rendered and style images, enabling nearest-neighbor matching and style loss computation to guide optimization.}
    \label{fig:main-workflow}
\end{figure*}

\subsection{Preliminaries}
\subsubsection{3D Stylization}\label{sec:notation}

\begin{table}[ht]
\caption{Summary of core notation used in our method.}
\centering
\begin{tabular}{ll}
\toprule
\textbf{Symbol} & \textbf{Meaning} \\
\midrule
$P_k$            & Camera pose of $k$-th view \\
$I_k$            & Input image from the $k$-th view \\
$J$              & Style reference image \\
$I_k^J$          & Stylized image matching style $J$ \\
$M$              & Reconstructed surface mesh \\
$\boldsymbol{x}$ & A 3D point in space \\ 
$\boldsymbol{r}$ & A ray in 3D space \\
$\boldsymbol{d}$ & Viewing direction \\
$\boldsymbol{v}_i$ & Coordinates of the $i$-th vertex \\
$\Delta \boldsymbol v_i$ & Position offset of vertex  $\boldsymbol{v}_i$ \\
$S$              & 3D surface \\
$[u, v]$         & Pixel location in image space \\
$f$              & View synthesizer (e.g., NeRF) \\
$g^J$            & Stylized view synthesizer matching style $J$ \\
$a(\boldsymbol{x})$ & Appearance feature at surface point $\boldsymbol{x}$ \\
$G$              & Grid-based appearance representation \\
$h^J(\cdot)$     & Image-level stylizer corresponding to style $J$ \\
$\mathcal{M}$    & Visibility mask \\ 
$\mathcal{T}_{l \to k}(\cdot, \cdot)$   & Transformation function from view $l$ to $k$ \\
$\mathcal{F}$    & Feature extracted by a VGG network \\
\bottomrule
\end{tabular}
\label{tab:notation}
\end{table}
The goal of 3D scene stylization is to modify the visual appearance of a scene to match the style of a reference image while preserving its underlying 3D geometry.
Formally, given a 3D scene representation $f$ and a style reference image $J$, the stylization task aims to learn a view synthesizer $g^J$ that can render the scene from arbitrary viewpoints in the desired style.
To facilitate the formulation of our method, we adopt the following notation. Boldface symbols (e.g., $\boldsymbol{x}$) denote vectors, and the core symbols used throughout the paper are summarized in Table~\ref{tab:notation}.

\subsubsection{Neural Radiance Fields}
A neural radiance field (NeRF) \cite{mildenhall2021nerf} is a learnable representation that enables the synthesis of novel views of a 3D scene.
Given a 3D point $\boldsymbol{x}$ and a viewing direction $\boldsymbol{d}$ corresponding to a camera ray, the model predicts a color $\boldsymbol{c}$ and a volumetric density $\sigma$, which together define the scene’s appearance along that ray. By accumulating these predictions along rays cast from a camera with given intrinsic and extrinsic parameters, NeRF can render an image from an arbitrary viewpoint. 
This ability to synthesize images from arbitrary viewpoints makes NeRF a natural foundation for 3D stylization. Specifically, it enables the formulation of how a scene should appear under a given style from any viewpoint. More importantly, because the rendering process is governed by a shared 3D representation, NeRF inherently enforces geometric consistency across views. This property can be leveraged to ensure that stylized renderings from different viewpoints remain coherent, which is a key requirement for consistent 3D scene stylization.

\subsection{Stylization via Explicit Surface Representation}
Surf-style takes as input a trained radiance field $f$ of a scene and a
reference style image $J$.
The NeRF model $f$ implicitly encodes the scene geometry and part of the appearance information (contents). 
The reference image $J$ provides the remaining appearance
information (the desired style).
The objective of Surf-style is to produce a stylized view synthesizer $g^J$, a
function that maps a camera pose $P_k$ to a geometrically consistent stylized
image:
\begin{align}
    g^J(P_k) = \tilde{I}_k^J.
    \label{eq:stylizer}
\end{align}

As a baseline, consider a na\"{i}ve stylization approach: let $h^J(\cdot)$ denote an
image-level stylizer that applies the visual style of $J$ to an input image, as in \cite{gatys2016image, zhu2017unpaired}. 
Composing $h^J$ with the scene's radiance field $f$ results in a stylized novel view synthesizer. Specifically,
$h^J\circ f$ produces a stylized image for each camera viewpoint by applying 2D stylization independently to each rendered view.

The na\"{i}ve approach struggles to maintain consistency across different viewpoints.
Consider a surface point $\boldsymbol{x} \in S$
projected to pixel coordinates $[u_k, v_k]$ and $[u_l, v_l]$ in two
views $P_k$ and $P_l$, respectively.
Since stylization is applied independently in each view, 
the resulting colors at these two pixels are influenced by their respective local image patches.
Consequently, the stylized appearances of the same 3D point may differ across views. 
In contrast, a surface-based stylization approach applies style directly to the 3D surface $S$, 
which is then projected to the image plane. This enables consistent stylization regardless of the viewing angle.

Surf-style reconstructs a surface mesh to enforce consistent stylization of $S$ across views. 
It then builds a stylized renderer by reparameterizing the scene’s appearance along the mesh.
Fig.~\ref{fig:main-workflow} illustrates the overall workflow.

\subsubsection{Geometry Extraction}
We adopt the geometry extraction process from NeRF2Mesh\cite{Tang2023NeRF2Mesh}, which converts a trained NeRF model $f$ into a triangle mesh $M = (V, F)$, where $V$ and $F$ denote the set of vertices and faces, respectively.
This process involves surface extraction via an iterative mesh refinement algorithm.
We reuse this process without modification, as our focus lies in the stylization stage.
Once extracted, the mesh is kept fixed throughout stylization to preserve the scene's geometry and ensure consistency across views.

In NeRF2Mesh, geometry and appearance are jointly optimized by rendering an image from the mesh-based scene representation. Given a mesh $M$ and appearance features $G$, a differentiable renderer \cite{laine2020modular} $R$ synthesizes an image from a given camera pose $P_k$, as expressed by:
\begin{align}
R(M, G, P_k) \to I_k.
\label{eq:GE_render}
\end{align}
The output image is compared against ground-truth training images to guide learning. We adopt this optimization process only for geometry extraction and discard the learned appearance $G$ during stylization.

\subsubsection{Stylization}
To enable style transfer that aligns with surface geometry, we adopt a surface-based representation of appearance features. Specifically, instead of storing color information in a 3D grid $G$, we attach learnable appearance features directly to the vertices of the reconstructed mesh. Each vertex is assigned a view-independent diffuse color feature $\boldsymbol{c}_d \in \mathbb{R}^3$ and a view-dependent specular feature $\boldsymbol{c}_s \in \mathbb{R}^3$.

Given a camera pose, the color at a surface point $\boldsymbol{x}$ is computed by a custom shader. The shader first performs barycentric interpolation of the appearance features associated with the triangle’s vertices $\{\boldsymbol{v}_i, \boldsymbol{v}_j, \boldsymbol{v}_k\}$, whose positions define the triangle containing $\boldsymbol{x}$. This yields the interpolated features $(\boldsymbol{c^x}_d, \boldsymbol{c^x}_s)$. It then decodes the specular component based on the viewing direction $\boldsymbol{d}$. The final color of $\boldsymbol{x}$ is obtained as:
\begin{align}
\boldsymbol c = \boldsymbol{c^x}_d + \text{Decoder}(\boldsymbol d, \boldsymbol{c^x}_s),
\label{eq:color_rgb}
\end{align}
where $\text{Decoder}(\cdot, \cdot)$ is a neural network adapted from \cite{Tang2023NeRF2Mesh} that models the view-dependent specular effect.

\begin{figure}[t]
 \centering
 \includegraphics[width=0.9\columnwidth]{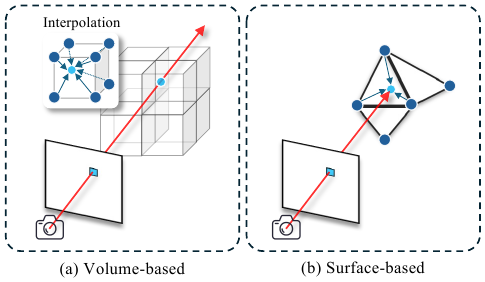}
 \caption{Volume-based rendering vs. Surface-based shading. In volume rendering, pixel colors are computed by interpolating features from surrounding grid points in 3D space, which often leads to imprecise and diffuse stylization. In contrast, our Surface-based approach stores appearance parameters directly on mesh vertices. A custom shader then interpolates these features along the mesh surface, enabling more spatially localized stylization effects.}
 \label{fig: vert_based_Renderer}
\end{figure}

In the original volume-based rendering for appearance representation $G$, each learnable parameter contributes visual information to an entire surrounding volume, which may not accurately reflect the surface structures and details. As shown in Fig. \ref{fig: vert_based_Renderer} (a), when a ray is cast from a pixel with a specific viewing direction, the point $\boldsymbol x$ intersecting the object surface is computed. Then the color at $\boldsymbol x$ is determined by interpolating the features on grid vertices surrounding $\boldsymbol x$. In this way, the changes to the grid parameters affect a broader region of the scene and result in imprecise editing.
In contrast, our surface-based approach confines learnable appearance information to well-defined surface elements. As shown in Fig.~\ref{fig: vert_based_Renderer} (b), this method localizes stylization effects to the mesh structure itself, enabling higher fidelity to surface geometry and detail.

During training, the geometry (mesh $M$) and the view-dependent specular components ($\{\boldsymbol c_s\}_{1 ... \left\vert V \right\vert}$) are obtained from the previous stage and kept fixed to preserve appearance consistency. The view-independent diffuse features ($\{\boldsymbol c_d\}_{1 ... \left\vert V \right\vert}$) are optimized to minimize a stylization loss, which we describe in the next subsection.

The final output of the stylization stage is the function $g^J$, comprising the mesh $M$ and its associated stylized per-vertex features.

\subsubsection{Loss} 
This section outlines the loss functions used in the three stages of the pipeline.

\textbf{Photometric Loss.}
During the training of the radiance field, we adopt the original NeRF rendering loss:

\begin{align}
\mathcal{L}_{\text{photo}} = \sum_{\boldsymbol r} \|c(\boldsymbol r) - \hat{c}(\boldsymbol r) \|^2,
\end{align}
where $c(\boldsymbol r)$ is the truth color and $\hat{c}(\boldsymbol r)$ is the predicted color for the ray $\boldsymbol r$.

\textbf{Geometric Loss.}
In the surface extraction and refinement stage, we apply a Laplacian smoothing loss function following \cite{Tang2023NeRF2Mesh}:

\begin{align}
\mathcal{L}_{\text{smooth}} = \sum_i \sum_{j \in s_i} \frac{1}{\lvert s_i \rvert} \|(\boldsymbol{v}_i + \Delta \boldsymbol{v}_i) - (\boldsymbol{v}_j + \Delta \boldsymbol{v}_j) \|^2,
\end{align}
where $s_i$ denotes the indices of the neighbors of vertex $\boldsymbol{v}_i$, and $\Delta \boldsymbol{v}_i$ is a trainable offset assigned to each vertex $\boldsymbol{v}_i$ in the geometry extraction stage, following \cite{Tang2023NeRF2Mesh}. Additionally, L2 regularization is applied for the vertices offset:

\begin{align}
\mathcal{L}_{\text{offset}} = \sum_i \|\Delta \boldsymbol v_i \|^2.
\end{align}

\textbf{Stylization Loss.}
For stylization, we adopt Nearest Neighbor Feature Matching (NNFM) loss following ARF \cite{zhang2022arf}: 

\begin{align}
\mathcal{L}_{\text{nnfm}} (\mathcal{F}_{\tilde I}, \mathcal{F}_J) = \frac{1}{N} \sum_{u,v} \min_{\hat{u}, \hat{v}} \mathcal{D} (\mathcal{F}_{\tilde I}(u, v), \mathcal{F}_J(\hat{u}, \hat{v})),
\label{eq:style_loss}
\end{align}
where $\mathcal{F}_{\tilde I}$ and $\mathcal{F}_J$ refer to VGG feature maps extracted from rendered image $\tilde I$ and style image $J$, respectively. $\mathcal{F}_{\tilde I}(u, v)$ represent the feature vector at pixel location $(u, v)$ of the feature map $\mathcal{F}_{\tilde I}$, and $N$ is the number of pixels in $\mathcal{F}_{\tilde I}$. 
The feature distance $\mathcal{D}(\boldsymbol a, \boldsymbol b)$ is computed using cosine distance:

\begin{align}
\mathcal{D}(\boldsymbol{a}, \boldsymbol{b}) = 1 - \frac{\boldsymbol{a}^\top \boldsymbol{b}}{\|\boldsymbol{a}\| \|\boldsymbol{b}\|}.
\end{align}

To control the strength of stylization, we also incorporate a content loss:
\begin{align}
\mathcal{L} = \mathcal{L}_{\text{nnfm}} (\mathcal{F}_{\tilde I}, \mathcal{F}_J) + \lambda \cdot \mathcal{L}_2(\mathcal{F}_{\tilde I}, \mathcal{F}_I),
\label{eq:style_loss_strength}
\end{align}
where $\lambda$ is the content weight, with a higher $\lambda$ emphasizing content preservation.  $\mathcal{F}$ denotes the VGG features extracted from the respective image.

\section{Experiments}
In this section, we report experiments that evaluate the effectiveness of the Surf-Style model in addressing the challenge of maintaining cross-view consistency in 3D stylization. 
To assess the geometric consistency of a stylized 3D scene, we utilize a geometric consistency cost metric that evaluates coherence in sequences of stylized renderings by leveraging the expected optical flow of subsequent frames.
The Surf-Style model also allows users to specify regions for applying a desired style and control the styles of different parts or objects separately within the scene.

\subsection{Datasets} 
We evaluated the performance of Surf-Style using two datasets commonly used in image-based reconstruction research.

\textbf{NeRF-Synthetic} \cite{mildenhall2021nerf} 
consists of eight synthetic 3D scenes, each containing a central object with clearly defined foreground masks and ground-truth camera poses. The dataset provides high-resolution images. And for our experiments, we use a resolution of 800 × 800 pixels per image. It also features well-categorized training and testing splits, allowing for structured evaluation of 3D scene reconstruction and stylization performance.

\textbf{NeRF-LLFF} \cite{mildenhall2019local} 
contains real-world 3D scenes captured from multiple viewpoints with known camera poses. Unlike the NeRF-Synthetic dataset, LLFF features more complex surfaces, lighting variations, and occlusions, making it well-suited for evaluating the robustness of stylization techniques in real-world scenarios. For our experiments, we use the following scenes: \textit{Fern}, \textit{Flower}, \textit{Horns}, \textit{Leaves}, and \textit{Room}. The dataset provides high-resolution images, which we downsample by a factor of 4, following the setup used in prior works \cite{zhang2023refnpr, Fridovich2022, Tang2023NeRF2Mesh}.

\subsection{Implementation Details} 
We use the outputs of the geometry extraction module (mesh and appearance field from NeRF2Mesh \cite{Tang2023NeRF2Mesh}) as input to our stylization process. Stylization is performed by optimizing per-vertex appearance features while keeping the mesh geometry fixed.

We adopt the NNFM loss from ARF \cite{zhang2022arf}, using a pre-trained VGG-16 network for feature extraction. Following ARF, we use features from the \textit{conv3} block in VGG-16 and set the content loss weight $\lambda$ in \eqref{eq:style_loss_strength} to 0.005. Each scene is trained for 1000 iterations.

We compare our model with the state-of-the-art methods: ARF \cite{zhang2022arf}, Ref-NPR \cite{zhang2023refnpr}, and SNeRF \cite{nguyen2022snerf}. While Ref-NPR manually creates reference images, in our experiments, we generate the reference image by applying a pre-trained 2D style transfer model \cite{gatys2016image} to the same style image used in our stylization module.

\begin{table*}
\centering
\caption{\label{tab:synthetic_result} Comparison of Geometric Consistency Cost (GCC) in the Stylization of the NeRF-Synthetic Dataset \cite{mildenhall2021nerf}. }
\begin{tabular}{l | c | c | c | c | c | c | c | c }
\hline
Method & Lego & Ship & Chair & Mic & Drums & Ficus & Hotdog & Materials \\\hline
ARF \cite{zhang2022arf} & 0.029 & 0.015 & 0.027 & 0.023 & 0.031 & 0.014 & 0.015 & 0.044 \\
Ref-NPR \cite{zhang2023refnpr} & 0.021 & 0.016 & 0.015 & 0.018 & 0.027 & \textbf{0.010} & 0.011 & 0.025\\
Surf-Style & \textbf{0.017} & \textbf{0.010} & \textbf{0.010} & \textbf{0.013} & \textbf{0.025} & 0.013 & \textbf{0.008} & \textbf{0.020} \\
\hline
\end{tabular}
\\
{\raggedright \footnotesize Since SNeRF \cite{nguyen2022snerf} has not reported the results on the NeRF-Synthetic dataset, we do not include it in this comparison. \par}
\end{table*}

\begin{table}
\centering
\caption{\label{tab:llff_result} Comparison of Geometric Consistency Cost (GCC) in the Stylization of the LLFF Dataset \cite{mildenhall2019local}.}
\begin{tabular}{l | c | c | c | c | c }
\hline
Method & Fern & Flower & Horns & Leaves & Room \\\hline
ARF \cite{zhang2022arf} & 0.015 & 0.018 & 0.020 & 0.020 & 0.016 \\
SNeRF \cite{nguyen2022snerf} & 0.011 & 0.010 & 0.011 & \textbf{0.009} & 0.011 \\
Ref-NPR \cite{zhang2023refnpr} & 0.010 & 0.012 & 0.014 & 0.013 & 0.010 \\
Surf-Style & \textbf{0.009} & \textbf{0.006} & \textbf{0.010} & 0.014 & \textbf{0.004} \\
\hline
\end{tabular}
\end{table}

\subsection{Geometric Consistency in Non-Photorealistic Stylization{\label{subsec:exper:main-results}}}

\subsubsection{Evaluation Metrics} 

To evaluate the geometric consistency, we adopt the concept of short-term temporal consistency, which is commonly used in video style transfer \cite{lai2018, ruder2016}, and follow the evaluation metrics used in \cite{nguyen2022snerf}.
Given a stylized novel view synthesizer $g$ defined in \eqref{eq:stylizer}, the geometric consistency cost(GCC) quantifies the consistency between consecutive stylized views rendered by $g$ under varying camera poses. Specifically, GCC is defined as,

\begin{align} \label{eq:geometry_consistency}
\text{GCC}( g ) = 
\sum_{k=2}^{m} 
\sum_{u, v} 
    \bigg(
        & \mathcal M_{k, k-1}[u,v] 
        \cdot 
        \\
        & \big\| 
            \tilde I_k [u, v] 
            - 
            \tilde I_{k-1} [ u', v' ]
        \big\|_2^2 
    \bigg) \nonumber,
\end{align}
where $\tilde I_k$ refers to an image rendered from camera pose $P_k$, and $[u, v]$ specify a pixel location.
The transformed pixel location $(u', v') = \mathcal{T}_{k \to k-1}(u, v)$ is obtained by applying the transformation that maps a pixel $[u, v]$ in frame $k$ to its corresponding location in frame $k{-}1$.
$\mathcal M_{k, k-1}(u, v)$ 
equal to 1 when pixel $[u, v]$ at $\tilde I_k$ can be found at $\tilde I_{k-1}$, otherwise, it is 0.
$m$ refers to the total rendering frames.

In practice, the common approach is to use optical flow models as transformation $\mathcal T_{k \to k-1}$.  Specifically, we adopt the pre-trained RAFT model \cite{teed2020raft} as used in \cite{nguyen2022snerf}.
Additionally, the visibility mask $\mathcal M_{k, k-1}$ is slightly different from the original definition in \eqref{eq:geometry_consistency}. We follow \cite{lai2018} and use a smoothed mask to enable a more robust evaluation of geometric consistency. It is calculated from the wrapping error between input frames $\tilde I_{k}$ and warped input frame $\hat{I}_{k-1}$,
\begin{align}
    \mathcal M_{k, k-1} = \exp(-\alpha \|\tilde I_{k} - \hat{I}_{k-1}\|^2_2),
    \label{eq:visibility_mask}
\end{align}
where the $\alpha$ is set to be 50.

To evaluate a model $g$, we rendered two videos using testing camera poses. One video is from the pre-trained radiance field $f$, $\{I^k\}_{k=1}^m$, which is used to compute the optical flow. The other one is from the stylized novel view synthesizer $g$ according to \eqref{eq:stylizer}, $\{\tilde I^k\}_{k=1}^m$. For each video, $m$ images are rendered utilizing the rendering poses in testing data.
The optical flow $\mathcal T_{k \to k-1}$ is computed on $I^k$ and $I^{k-1}$ by RAFT \cite{teed2020raft}. Then it warps $\tilde I^{k-1}$ into $\hat I^{k-1}$.

As a result, the GCC is implemented in practice as follows,
\begin{align}
\text{GCC}(g^J) = \sum_{\mathclap{k=1}}^{m} \frac{1}{m} \mathcal M_{k, k-1} \|\tilde I^k - \hat I^{k-1}\|^2_2.
\label{eq:final_error}
\end{align}

\subsubsection{Results}

\textbf{Quantitative Analysis. }
Using the metrics described above, we evaluate the geometric consistency of stylized models generated by Surf-Style and the state-of-the-art models: ARF \cite{zhang2022arf}, Ref-NPR \cite{zhang2023refnpr}, and SNeRF \cite{nguyen2022snerf}. Table.~\ref{tab:synthetic_result} and Tab.~\ref{tab:llff_result} present the per-scene results on the NeRF-Synthetic \cite{mildenhall2021nerf} and LLFF \cite{mildenhall2019local} datasets, respectively.

The results demonstrate that Surf-Style achieves superior geometric consistency, as evidenced by lower geometric consistency cost across most scenes. However, in a few cases (i.e., \textit{Ficus}, \textit{Leaves}), our performance is on par with other methods. This limitation is mainly attributed to the quality of the mesh, which could be enhanced using more advanced mesh extraction techniques. Nevertheless, this aspect falls outside the primary scope of this study. 

\begin{figure*}[!ht]
    \centering
    \includegraphics[width=\textwidth]{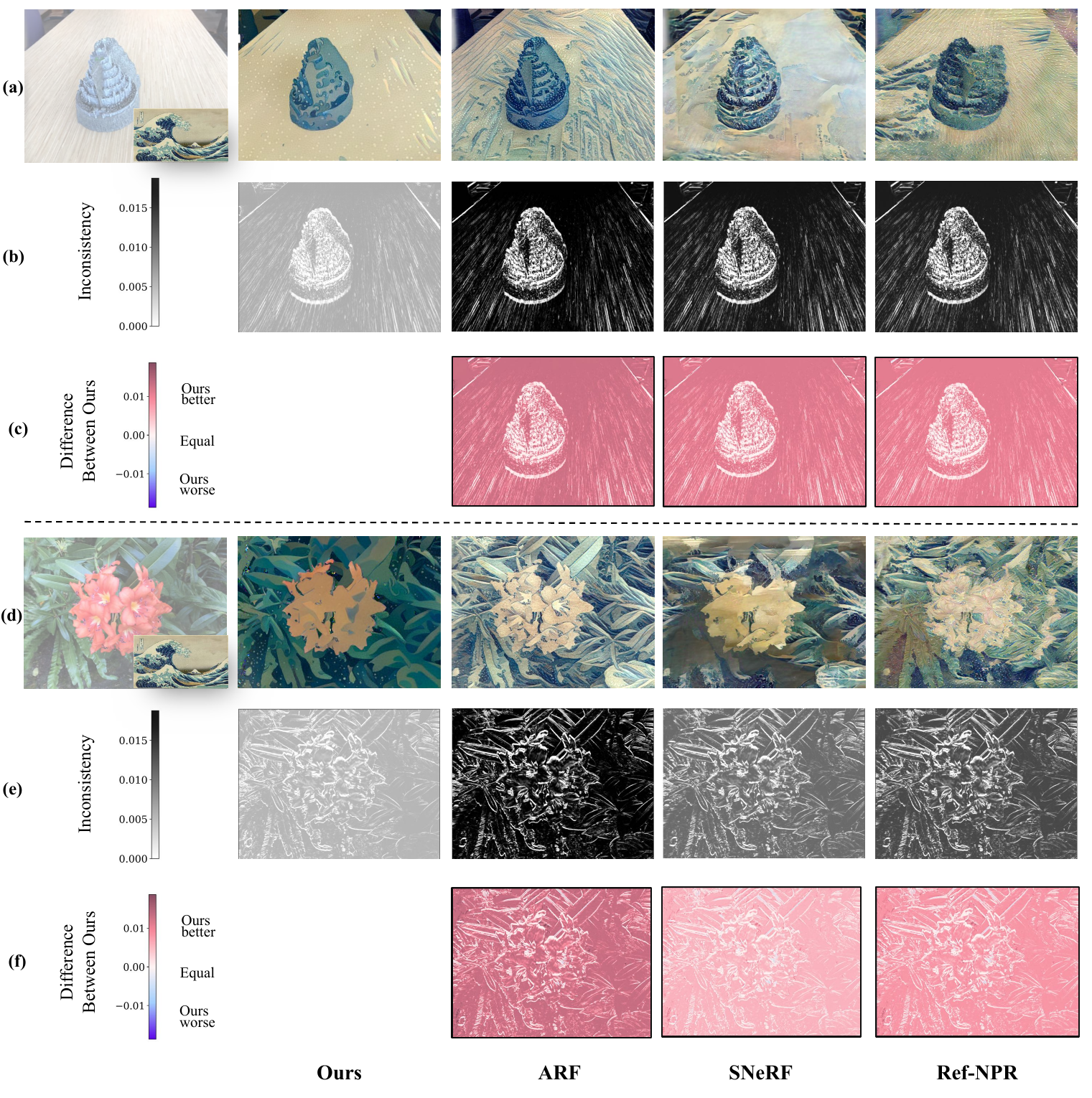}
    \caption{Visual comparison of geometric consistency cost.
The figure compares geometric consistency cost across two scenes: \textit{Fortress} (top) and \textit{Flower} (bottom). Row \textbf{(a)} and \textbf{(d)} show stylized results, with the content scene and style displayed in the first column. 
Rows \textbf{(b)} and \textbf{(e)} visualize geometric inconsistency by warping the previous frame $\tilde{I}^{k-1}$ to the current view using optical flow, producing $\hat{I}^{k-1}$. The inconsistency map is computed as the visibility-weighted difference between $\hat{I}^{k-1}$ and $\tilde{I}^k$ (see \eqref{eq:visibility_mask}).
Rows \textbf{(c)} and \textbf{(f)} compare these inconsistency maps across methods (ARF, SNeRF, Ref-NPR, and Ours), showing that our approach yields significantly lower inconsistency, especially in planar regions. }
    \label{fig:error_map}
\end{figure*}

\textbf{Qualitative Analysis. }
Fig. \ref{fig:error_map} qualitatively illustrates the geometric consistency cost evaluated for different models, using pairs of adjacent frames to generate inconsistency maps based on~\eqref{eq:final_error}. Furthermore, the difference between these inconsistency maps is computed to assess the performance of Surf-Style across various 3D regions. The results highlight that Surf-Style demonstrates significant advantages in planar regions, where other approaches tend to distort scene geometry to align with the target style. These findings reinforce the importance of this work in preserving geometric consistency.

\textbf{Dataset-Specific Analysis.}
In the NeRF-Synthetic dataset, each scene contains a single object centered within the 3D space, with no background. This unique structure poses a challenge for stylization techniques, as the absence of background requires stylization to be constrained to the object’s surface rather than applied across the entire 3D field. However, existing methods \cite{huang2022stylizednerf, chiang2022stylizing, nguyen2022snerf} are not well-suited for this dataset because they rely on stylization losses computed from features extracted across the entire field, and it is not straightforward for them to locate the surface, leading to undesired style transfer onto background areas.

To overcome this limitation, Surf-Style leverages geometric information by utilizing a mesh representation of the surface of the object. This allows us to identify which pixels in the synthesized image correspond to the surface and which do not. During training, our stylized synthesizer generates an image, and for pixels where the rendering ray hits the surface, the stylized output reflects the object's features. For the remaining pixels (those not hitting the surface), we fill in the corresponding regions with the style image. This ensures that features extracted from non-surface pixels remain identical to those in the target style image, reducing their impact on the training process. Consequently, the model is encouraged to update color parameters only at surface vertices, effectively constraining the stylization to the object's surface and minimizing background interference.

\begin{figure*}[h!]
    \centering
    \includegraphics[width=0.9\textwidth]{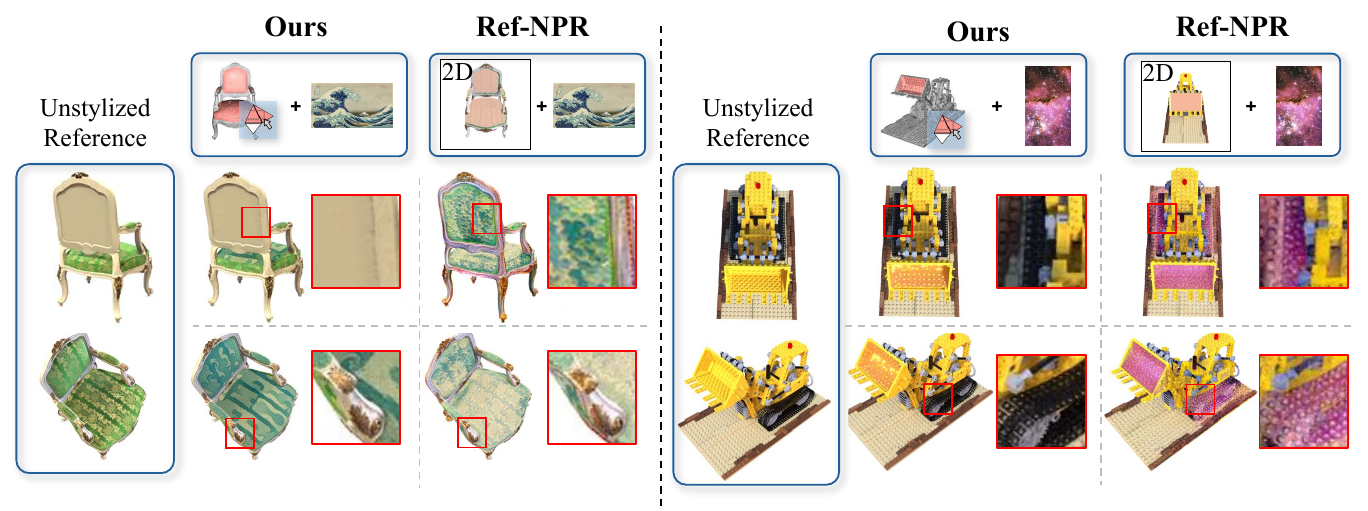}
    \caption{Demonstration of the localized control achieved by Surf-Style model. This figure compares Surf-Style with Ref-NPR. In our approach, target areas such as the chair's cushion in scene \textit{Chair} and the inner surface of the bucket in scene \textit{LEGO} are selected using a 3D mesh editor. In contrast, Ref-NPR only supports selection through 2D masks. Our method restricts stylization to the desired surfaces, while Ref-NPR lacks precision, spilling stylization onto unintended areas, i.e., the cushion and the back of the chair in the \textit{Chair} scene, and the wheels in the \textit{LEGO} scene.}
    \label{fig:area-controllable_comparison}
\end{figure*}

\begin{figure*}[h]
    \centering
    \includegraphics[width=0.8\textwidth]{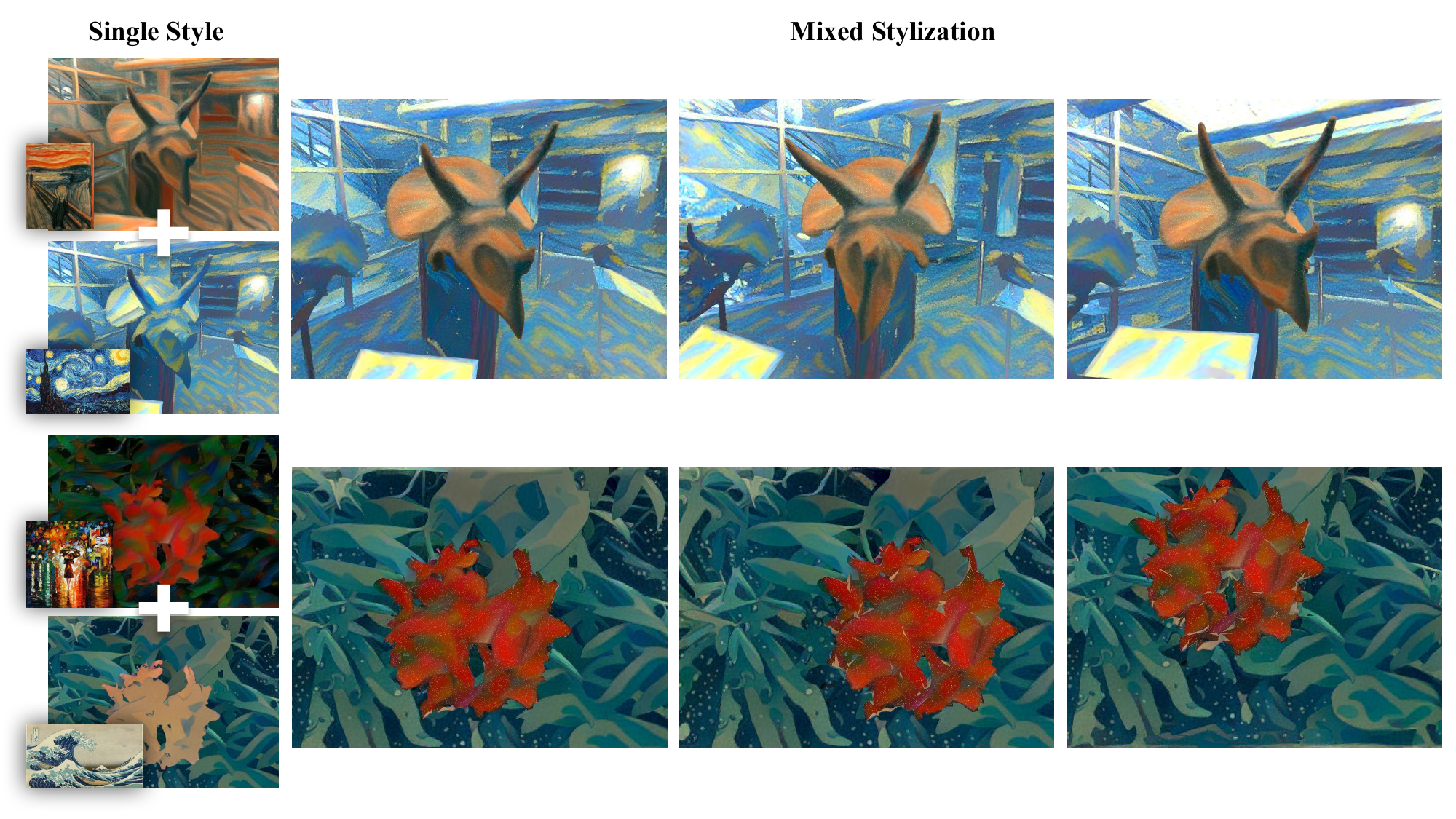}
    \caption{Illustration of Surf-Style's flexibility in combining multiple styles within a scene. The left side displays single-style results alongside their corresponding style images. The right side shows a mixed stylization result, where the area of interest is stylized using the top style, and the remaining region adopts the lower style. This demonstrates our model's capability to blend multiple styles within a scene.}
    \label{fig:exp_example2}
\end{figure*}

\subsection{Advantages of Mesh-Based Representation for Stylization Control}

In this section, we address the task of area-controllable stylization, showing the advantages of Surf-Style by conducting qualitative comparisons. 
Overall, the Surf-Style model allows users to specify physically meaningful regions for applying a desired style and independently edit the style of distinct object parts or different objects within the scene.
We present two experiments to demonstrate our model's unique capability of achieving localized stylization, as well as its flexibility in applying multiple styles to different parts of a scene.

\subsubsection{Localized Stylization} 
This experiment highlights the unique advantage of our surface-based framework, which enables stylization only in expected areas. As shown in Fig. \ref{fig:area-controllable_comparison}, we use an existing interactive mesh editor to select specific faces of interest (e.g., the cushion of a chair, the inner surface of the bucket in the LEGO shovel loader) and apply stylization exclusively to these faces. By explicitly storing scene appearance information on the mesh vertices, it becomes intuitive to localize and modify the visual style of the selected region.

In contrast, existing volumetric approaches, such as Ref-NPR, lack convenient region selection capabilities and often inadvertently affect surrounding regions. Fig. \ref{fig:area-controllable_comparison} presents controlled stylization results produced by Ref-NPR,  where a 2D mask is used to select 3D areas, but it unintentionally stylizes broader regions (e.g., the back of a chair and the wheel of the LEGO). The impact of different representations on region selection is illustrated in \ref{fig:volume_surface_compare}. This limitation arises from the reliance of Ref-NPR on rough estimates based on its volumetric density field, which lacks the precision of our explicit surface modeling. The issue is even more pronounced in approaches like ARF and SNeRF, which only support basic volume selection, typically relying on manually estimated 3D boxes.

\subsubsection{Multi-Region Stylization} 
This experiment showcases the versatility of Surf-Style in applying multiple styles to a single 3D scene. As shown in Fig. \ref{fig:exp_example2}, we demonstrate this using the scene \textit{Flower}, where the foreground (i.e., the flower) is stylized with \textit{Rain Princess}, while the background adopts \textit{The Great Wave off Kanagawa}. The results highlight our framework’s ability to integrate multiple styles within a scene without interference. This is achieved through surface-based appearance modeling, which leverages mesh faces as distinct units to associate object elements with visual information. Notably, our framework supports the use of more than two styles.

\section{Conclusion}
We have introduced a scene stylization framework that leverages explicit surface modeling to enhance cross-view consistency and controllability. 
Specifically, we reconstruct a mesh as the explicit geometric representation and store appearance information on its vertices, ensuring consistent rendering across different viewpoints. 
We support localized stylization by attaching view-independent and view-dependent appearance features to mesh vertices, allowing flexible control over the style applied to different regions of the surface.
Experimental results demonstrate that our method outperforms existing approaches in both qualitative and quantitative evaluations of consistency. Moreover, we highlight the capability of our framework to apply distinct styles to specific regions of a scene while maintaining overall consistency across viewpoints.

\section*{Acknowledgments}
The authors acknowledge the use of artificial intelligence–assisted tools for English grammar checking and language refinement in the preparation of this manuscript.


\begin{thebibliography}{00}
\bibitem{Sun22} C. Sun, M. Sun, and H.-T. Chen, ``Direct voxel grid optimization: Super-fast convergence for radiance fields reconstruction,'' in \emph{Proc. CVPR}, New York, USA, 2022, pp. 5449--5459.

\bibitem{Xiang2021NeuTexNT} F. Xiang, Z. Xu, M. Havsan, Y. Hold-Geoffroy, K. Sunkavalli, and H. Su, ``Neutex: Neural texture mapping for volumetric neural rendering,'' in \emph{Proc. CVPR}, New York, USA, 2021, pp. 7115--7124.

\bibitem{Chen2022MobileNeRF} Z. Chen, T. A. Funkhouser, P. Hedman, and A. Tagliasacchi, ``Mobilenerf: Exploiting the polygon rasterization pipeline for efficient neural field rendering on mobile architectures,'' in \emph{Proc. CVPR}, New York, USA, 2023, pp. 16569--16578.

\bibitem{isola2017image} P. Isola, J.-Y. Zhu, T. Zhou, and A. A. Efros, ``Image-to-image translation with conditional adversarial networks,'' in \emph{Proc. CVPR}, New York, USA, 2017, pp. 1125--1134.

\bibitem{hollein2022stylemesh} L. H{\"o}llein, J. Johnson, and M. Nie{\ss}ner, ``Stylemesh: Style transfer for indoor 3d scene reconstructions,'' in \emph{Proc. CVPR}, New York, USA, 2022, pp. 6198--6208.

\bibitem{zhang2023refnpr} Y. Zhang, Z. He, J. Xing, X. Yao, and J. Jia, ``Ref-npr: Reference-based non-photorealistic radiance fields for controllable scene stylization,'' in \emph{Proc. CVPR}, New York, USA, 2023, pp. 4242--4251.

\bibitem{zhang2018lpips} R. Zhang, P. Isola, A. A. Efros, E. Shechtman, and O. Wang, ``The unreasonable effectiveness of deep features as a perceptual metric,'' in \emph{Proc. CVPR}, New York, USA, 2018, pp. 586--595.

\bibitem{huang2022stylizednerf} Y. Huang, Y. He, Y. Yuan, Y. Lai, and L. Gao, ``Stylizednerf: Consistent 3d scene stylization as stylized nerf via 2d-3d mutual learning,'' in \emph{Proc. CVPR}, New York, USA, 2022, pp. 18342--18352.

\bibitem{Sun2018PWC-Net} D. Sun, X. Yang, M.-Y. Liu, and J. Kautz, ``Pwc-net: Cnns for optical flow using pyramid, warping, and cost volume,'' in \emph{Proc. CVPR}, New York, USA, 2018, pp. 8934--8943.

\bibitem{Ilg2017} E. Ilg, N. Mayer, T. Saikia, M. Keuper, A. Dosovitskiy, and T. Brox, ``Flownet 2.0: Evolution of optical flow estimation with deep networks,'' in \emph{Proc. CVPR}, New York, USA, 2017, pp. 1647--1655.

\bibitem{Fridovich2022} S. Fridovich-Keil, A. Yu, M. Tancik, Q. Chen, B. Recht, and A. Kanazawa, ``Plenoxels: Radiance fields without neural networks,'' in \emph{Proc. CVPR}, New York, USA, 2022, pp. 5491--5500.

\bibitem{yuan2022editing} Y.-J. Yuan, Y.-T. Sun, Y.-K. Lai, Y. Ma, R. Jia, and L. Gao, ``Nerf-editing: Geometry editing of neural radiance fields,'' in \emph{Proc. CVPR}, New York, USA, 2022, pp. 18353--18364.

\bibitem{pumarola2021dnerf} A. Pumarola, E. Corona, G. Pons-Moll, and F. Moreno-Noguer, ``D-nerf: Neural radiance fields for dynamic scenes,'' in \emph{Proc. CVPR}, New York, USA, 2021, pp. 10318--10327.

\bibitem{Barron22_Mip360} J. T. Barron, B. Mildenhall, D. Verbin, P. P. Srinivasan, and P. Hedman, ``Mip-nerf 360: Unbounded anti-aliased neural radiance fields,'' in \emph{Proc. CVPR}, New York, USA, 2022, pp. 5460--5469.

\bibitem{zhu2017unpaired} J.-Y. Zhu, T. Park, P. Isola, and A. A. Efros, ``Unpaired image-to-image translation using cycle-consistent adversarial networks,'' in \emph{Proc. CVPR}, New York, USA, 2017, pp. 2223--2232.

\bibitem{gatys2016image} L. A. Gatys, A. S. Ecker, and M. Bethge, ``Image style transfer using convolutional neural networks,'' in \emph{Proc. CVPR}, New York, USA, 2016, pp. 2414--2423.

\bibitem{radl2023laenerf} L. Radl, M. Steiner, A. Kurz, and M. Steinberger, ``Laenerf: Local appearance editing for neural radiance fields,'' in \emph{Proc. CVPR}, New York, USA, 2024, pp. 4969--4978.

\bibitem{lai2018} W.-S. Lai, J.-B. Huang, O. Wang, E. Shechtman, E. Yumer, and M.-H. Yang, ``Learning blind video temporal consistency,'' in \emph{Proc. ECCV}, 2018, pp. 170--185.

\bibitem{Barron_mipnerf} J. T. Barron, B. Mildenhall, M. Tancik, P. Hedman, R. Martin-Brualla, and P. P. Srinivasan, ``Mip-nerf: A multiscale representation for anti-aliasing neural radiance fields,'' in \emph{Proc. ICCV}, 2021, pp. 5835--5844.

\bibitem{saroha2024gsstyle} A. Saroha, M. Gladkova, C. Curreli, D. Muhle, T. Yenamandra, and D. Cremers, ``Gaussian splatting in style,'' in \emph{Proc. GCPR}, 2024, pp. 234--251.

\bibitem{liu2024stylegaussian} K. Liu, F. Zhan, M. Xu, C. Theobalt, L. Shao, and S. Lu, ``Stylegaussian: Instant 3d style transfer with gaussian splatting,'' in \emph{Proc. SIGGRAPH Asia}, 2024, pp. 1--4.

\bibitem{goel2022styletrf} R. Goel, D. Sirikonda, S. Saini, and P. J. Narayanan, ``Styletrf: Stylizing tensorial radiance fields,'' in \emph{Proc. ICVGIP}, 2022, pp. 1--9.

\bibitem{li2024instantstyle} S. Li and Y. Pan, ``Instant photorealistic neural radiance fields stylization,'' in \emph{Proc. ICASSP}, 2024, pp. 2980--2984.

\bibitem{kolkin2022nnst} N. Kolkin, M. Kucera, S. Paris, D. Sykora, E. Shechtman, and G. Shakhnarovich, ``Neural neighbor style transfer,'' in \emph{Proc. Eurographics}, 2022, pp. 25--28.

\bibitem{Wei2023NeuManifoldNW} X. Wei, F. Xiang, S. Bi, A. Chen, K. Sunkavalli, Z. Xu, and H. Su, ``Neumanifold: Neural watertight manifold reconstruction with efficient and high-quality rendering support,'' in \emph{Proc. WACV}, 2025, pp. 731--741.

\bibitem{Shen2021DMTet} T. Shen, J. Gao, K. Yin, M.-Y. Liu, and S. Fidler, ``Deep marching tetrahedra: A hybrid representation for high-resolution 3d shape synthesis,'' in \emph{Proc. NeurIPS}, Red Hook, NY, USA, 2021, pp. 6087--6101.

\bibitem{Wang2021} P. Wang, L. Liu, Y. Liu, C. Theobalt, T. Komura, and W. Wang, ``Neus: Learning neural implicit surfaces by volume rendering for multi-view reconstruction,'' in \emph{Proc. NeurIPS}, Red Hook, NY, USA, 2021, pp. 27171--27183.

\bibitem{Yariv2021VolSDF} L. Yariv, J. Gu, Y. Kasten, and Y. Lipman, ``Volume rendering of neural implicit surfaces,'' in \emph{Proc. NeurIPS}, Red Hook, NY, USA, 2021, pp. 4805--4815.

\bibitem{Liu22_DeVRF} J.-W. Liu et al., ``Devrf: Fast deformable voxel radiance fields for dynamic scenes,'' in \emph{Proc. NeurIPS}, Red Hook, NY, USA, 2022, pp. 36762--36775.

\bibitem{Oechsle2021UNISURF} M. Oechsle, S. Peng, and A. Geiger, ``Unisurf: Unifying neural implicit surfaces and radiance fields for multi-view reconstruction,'' in \emph{Proc. ICCV}, New York, USA, 2021, pp. 5569--5579.

\bibitem{Tang2023NeRF2Mesh} J. Tang et al., ``Delicate textured mesh recovery from nerf via adaptive surface refinement,'' in \emph{Proc. ICCV}, New York, USA, 2023, pp. 17693--17703.

\bibitem{zhang2022arf} K. Zhang et al., ``Arf: Artistic radiance fields,'' in \emph{Proc. ECCV}, Berlin, Heidelberg, 2022, pp. 717--733.

\bibitem{Yang2022NeuMesh} B. Yang et al., ``Neumesh: Learning disentangled neural mesh-based implicit field for geometry and texture editing,'' in \emph{Proc. ECCV}, Berlin, Heidelberg, 2022, pp. 597--614.

\bibitem{teed2020raft} Z. Teed and J. Deng, ``Raft: Recurrent all-pairs field transforms for optical flow,'' in \emph{Proc. ECCV}, Berlin, Heidelberg, 2020, pp. 402--419.

\bibitem{Chen22TensoRF} A. Chen et al., ``Tensorf: Tensorial radiance fields,'' in \emph{Proc. ECCV}, Berlin, Heidelberg, 2022, pp. 333--350.

\bibitem{cao2020psnet} X. Cao et al., ``Psnet: A style transfer network for point cloud stylization on geometry and color,'' in \emph{Proc. WACV}, New York, USA, 2020, pp. 3337--3345.

\bibitem{chiang2022stylizing} P.-Z. Chiang et al., ``Stylizing 3d scene via implicit representation and hypernetwork,'' in \emph{Proc. WACV}, New York, USA, 2022, pp. 1475--1484.

\bibitem{cheng2023AGAP} K. L. Cheng, Q. Wang, Z. Shi, K. Zheng, Y. Xu, H. Ouyang, Q. Chen, and Y. Shen, ``Learning naturally aggregated appearance for efficient 3d editing,'' in \emph{Proc. 3DV}, 2025, pp. 758--768.

\bibitem{zhang2024stylizedgs} D. Zhang, Y.-J. Yuan, Z. Chen, F.-L. Zhang, Z. He, S. Shan, and L. Gao, ``Stylizedgs: Controllable stylization for 3d gaussian splatting,'' \emph{IEEE Trans. Pattern Anal. Mach. Intell.}, vol. 47, no. 12, pp. 11961--11973, Dec. 2025, DOI: 10.1109/TPAMI.2025.3604010.

\bibitem{ruder2016} M. Ruder, A. Dosovitskiy, and T. Brox, ``Artistic style transfer for videos and spherical images,'' \emph{Int. J. Comput. Vis.}, vol. 126, pp. 1199--1219, Aug. 2018, DOI: 10.1007/s11263-018-1089-z.

\bibitem{chen2024upst} Y. Chen, Q. Yuan, Z. Li, Y. Liu, W. Wang, C. Xie, X. Wen, and Q. Yu, ``Upst-nerf: Universal photorealistic style transfer of neural radiance fields for 3d scene,'' \emph{IEEE Trans. Vis. Comput. Graph.}, vol. 31, pp. 2045--2057, Mar. 2024, DOI: 10.1109/TVCG.2024.3378692.

\bibitem{laine2020modular} S. Laine, J. Hellsten, T. Karras, Y. Seol, J. Lehtinen, and T. Aila, ``Modular primitives for high-performance differentiable rendering,'' \emph{ACM Trans. Graph.}, vol. 39, no. 6, pp. 1--14, Nov. 2020, DOI: 10.1145/3414685.3417861.

\bibitem{liu2018paparazzi} H.-T. D. Liu, M. Tao, and A. Jacobson, ``Paparazzi: Surface editing by way of multi-view image processing,'' \emph{ACM Trans. Graph.}, vol. 37, no. 6, pp. 1--11, Dec. 2018, DOI: 10.1145/3272127.3275047.

\bibitem{mildenhall2021nerf} B. Mildenhall, P. P. Srinivasan, M. Tancik, J. T. Barron, R. Ramamoorthi, and R. Ng, ``Nerf: Representing scenes as neural radiance fields for view synthesis,'' \emph{Commun. ACM}, vol. 65, no. 1, pp. 99--106, Dec. 2021, DOI: 10.1145/3503250.

\bibitem{mildenhall2019local} B. Mildenhall, P. P. Srinivasan, R. Ortiz-Cayon, N. K. Kalantari, R. Ramamoorthi, R. Ng, and A. Kar, ``Local light field fusion: Practical view synthesis with prescriptive sampling guidelines,'' \emph{ACM Trans. Graph.}, vol. 38, no. 4, pp. 1--14, Jun. 2019, DOI: 10.1145/3306346.3322980.

\bibitem{muller2022instant} T. M{\"u}ller, A. Evans, C. Schied, and A. Keller, ``Instant neural graphics primitives with a multiresolution hash encoding,'' \emph{ACM Trans. Graph.}, vol. 41, no. 4, pp. 1--15, Jul. 2022, DOI: 10.1145/3528223.3530127.

\bibitem{kerbl20233dgs} B. Kerbl, G. Kopanas, T. Leimk{\"u}hler, and G. Drettakis, ``3d gaussian splatting for real-time radiance field rendering,'' \emph{ACM Trans. Graph.}, vol. 42, no. 4, pp. 1--14, Jul. 2023, DOI: 10.1145/3592433.

\bibitem{nguyen2022snerf} T. Nguyen-Phuoc, F. Liu, and L. Xiao, ``Snerf: Stylized neural implicit representations for 3d scenes,'' \emph{ACM Trans. Graph.}, vol. 41, no. 4, pp. 1--11, Jul. 2022, DOI: 10.1145/3528223.3530107.

\bibitem{StylizedGS} D. Zhang, Y.-J. Yuan, Z. Chen, F.-L. Zhang, Z. He, S. Shan, and L. Gao, ``StylizedGS: Controllable Stylization for 3D Gaussian Splatting,'' \emph{IEEE Trans. Pattern Anal. Mach. Intell.}, vol. 47, no. 12, pp. 11961--11973, 2025.

\bibitem{Lin2021} Z. Lin, Z. Wang, H. Chen, X. Ma, C. Xie, W. Xing, L. Zhao, and W. Song, ``Image Style Transfer Algorithm Based on Semantic Segmentation,'' \emph{IEEE Access}, vol. 9, pp. 54518--54529, 2021.

\bibitem{Lee2020}
H. Lee and H.-C. Choi, ``Artifact-Free Image Style Transfer by Using Feature Map Clamping,'' \emph{IEEE Access}, vol. 8, pp. 89489--89496, 2020.


































\end{thebibliography}
\end{document}